\documentclass[10pt,twocolumn,letterpaper]{article}

\usepackage{iccv}
\usepackage{times}
\usepackage{epsfig}
\usepackage{graphicx}
\usepackage{amsmath}
\usepackage{amssymb}

\usepackage[breaklinks=true,bookmarks=false]{hyperref}

\iccvfinalcopy 

\def\iccvPaperID{****} 

\ificcvfinal\fi

\begin{document}

\title{A Global-Local Emebdding Module for Fashion Landmark Detection}

\author{Sumin Lee$^{1}$ \qquad \qquad  Sungchan Oh$^{2}$ \qquad  \qquad Chanho Jung$^{3}$ \qquad \qquad Changick Kim$^{1}$\\
$^{1}$Korea Advanced Institute of Science and Technology(KAIST)\\
$^{2}$Electronics and Telecommunications Research Institute\\
$^{3}$Hanbat National University\\
{\tt\small \{shum\_ming, changick\}@kaist.ac.kr, sungchan.oh@etri.re.kr, peterjung@hanbat.ac.kr}
}

\maketitle
\ificcvfinal\thispagestyle{empty}\fi

\begin{abstract}
   Detecting fashion landmarks is a fundamental technique for visual clothing analysis.
   Due to the large variation and non-rigid deformation of clothes,
   localizing fashion landmarks suffers from large spatial variances across poses, scales, and styles.
   Therefore, understanding contextual knowledge of clothes is required for accurate landmark detection.
   To that end, in this paper, we propose a fashion landmark detection network with a global-local embedding module.
   The global-local embedding module is based on a non-local operation for capturing long-range dependencies and a subsequent convolution operation for adopting local neighborhood relations.
   With this processing, the network can consider both global and local contextual knowledge for a clothing image.
   We demonstrate that our proposed method has an excellent ability to learn advanced deep feature representations for fashion landmark detection.
   Experimental results on two benchmark datasets show that the proposed network outperforms the state-of-the-art methods.
  Our code is available at \url{https://github.com/shumming/GLE_FLD}.
\end{abstract}

\section{Introduction}
\begin{figure}
\begin{center}
\includegraphics[width=83mm]{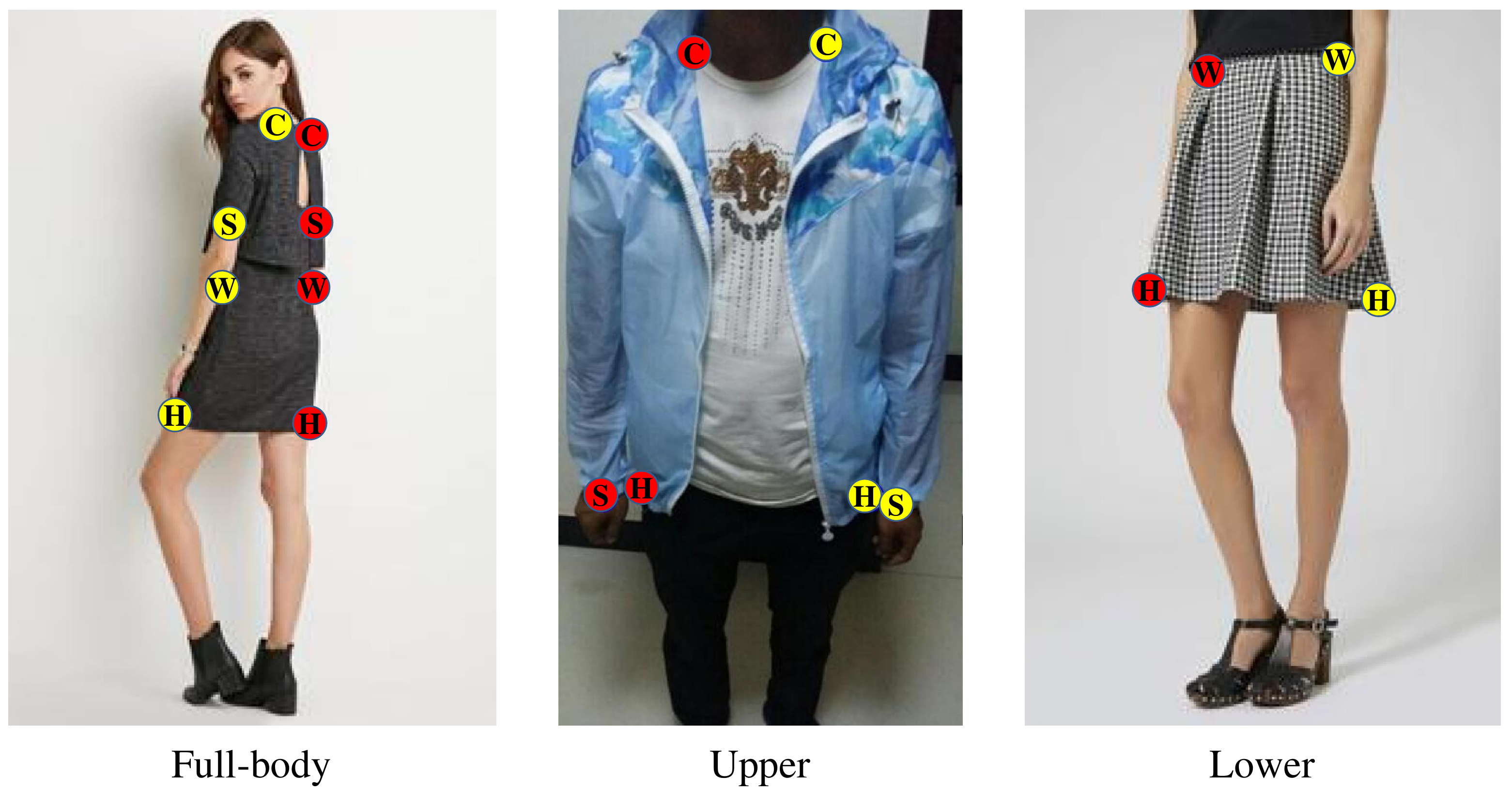}
\end{center}
  \caption{Examples of fashion landmarks. The landmarks represent collars (C), sleeves (S), waistlines (W), and hemlines (H). The red and yellow circles are used to indicate right and left sides in a image. Note that different types of clothes have different number of landmarks.}
\label{fig:landmark} 
\vspace{3mm}
\end{figure}

Visual fashion analysis has attracted research attention in recent years because of its huge potential usefulness in  industry.
With a development of large fashion datasets~\cite{DF1, FLD}, deep learning-based methods have achieved significant progresses in many tasks, such as clothes classification~\cite{DF1, BCRNN, ECCVW, DF2}, clothing retrieval~\cite{WTBI, DF1, DF2, retrieval}, and clothes generation~\cite{generation, generation2}.
The fashion analysis is a challenging task due to the large variation and non-rigid deformation of clothes in images.
To deal with this issue, recent methods~\cite{DF1, BCRNN, ECCVW, DF2} utilized fashion landmarks.
As illustrated in Fig~\ref{fig:landmark}, the fashion landmarks are key-points describing clothing structures such as collars, sleeves, waistlines, and hemlines.
Since these methods improved their performance considerably, fashion landmark detection is one of the key issues.
For localizing fashion landmarks, Liu et al.~\cite{FLD, DF1} utilized a regression method.
Wang et al.~\cite{BCRNN} designed a model to capture kinematic and symmetry grammar of clothing landmarks.
However, the fashion landmarks exhibit large spatial variances across poses, scales, and styles of clothing items.
Due to this property, the models require to understand comprehensive semantic information of clothes for accurate landmark detection.

To address this issue, we propose a fashion landmark detection network with a global-local embedding module to exploit rich contextual knowledge for clothes.
The global-local embedding module is specifically designed for embedding fashion landmark information by employing a non-local block~\cite{NLN} followed by convolutions.
Note that, in our proposed module, the non-local operation captures long-range global dependencies within a clothing image, and the following convolution operation enhances the local representation power of output features.
\begin{figure*}
\begin{center}
\includegraphics[width=170mm]{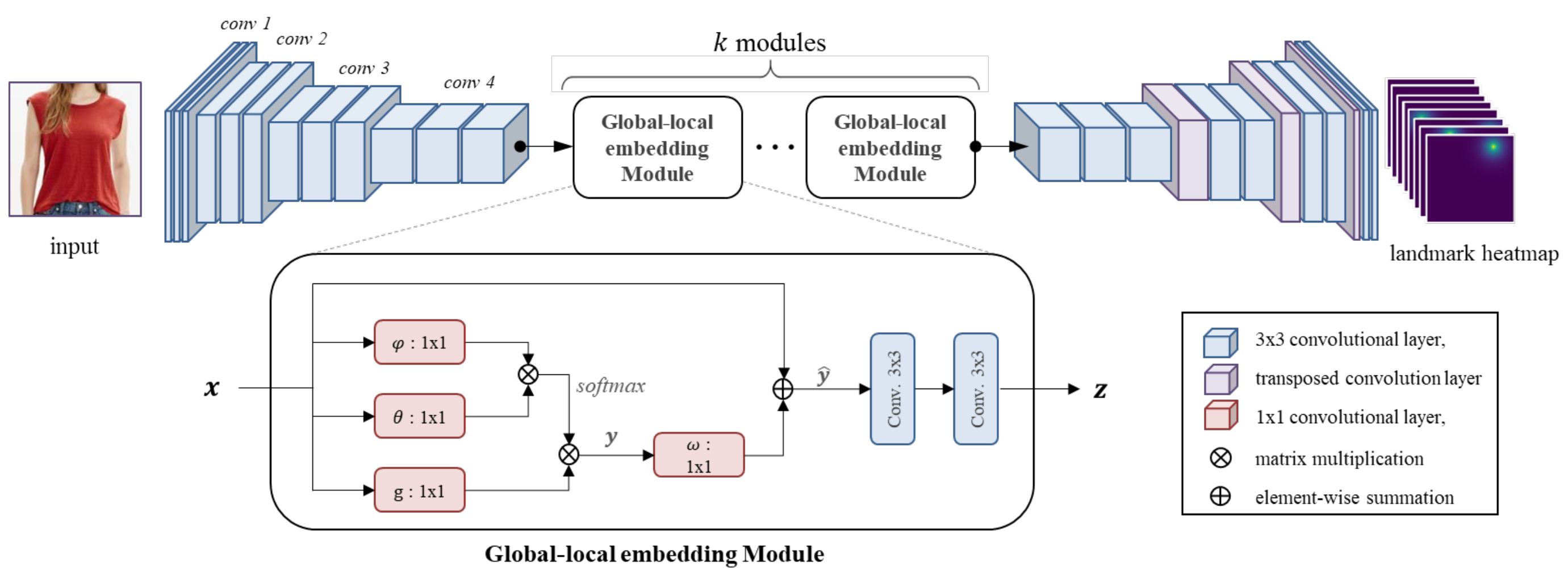}
\end{center}
  \caption{Illustration of our proposed network. Our network consists of basic convolutional network VGG-16 for feature extraction, the global-local embedding module for extracting rich landmark embedding features, and the upsampling network for predicting the landmark heatmaps.}
\label{fig:architecture}
\vspace{3mm}
\end{figure*}
With this processing, the output features are well facilitated to have global as well as local contextual information for input clothing image.
Lastly, by upsampling the feature map to the same size of the input fashion image, our network predicts high-resolution heatmaps for more accurate landmark localization.
To demonstrate the strength of our proposed network, we conduct experiments by using two datasets, Deepfashion~\cite{DF1} and FLD~\cite{FLD}.
Experimental results show that our network outperforms the state-of-the-art methods.

\begin{table*}[t]
\begin{center}
{\fontsize{9}{10}\selectfont
\renewcommand{\arraystretch}{1.2}
\begin{tabular}{cccccccccc}

\hline
\multicolumn{10}{c}{\textbf{FLD}}\\ \hline
\multicolumn{1}{c||}{}           & \multicolumn{1}{c}{L.Collar}        & \multicolumn{1}{c}{R.Collar}        & \multicolumn{1}{c}{L.Sleeve}        & \multicolumn{1}{c}{R.Sleeve}        & \multicolumn{1}{c}{L.Waistline}     & \multicolumn{1}{c}{R.Waistline}     & \multicolumn{1}{c}{L.Hem}           & \multicolumn{1}{c}{R.Hem}           & Avg.            \\ \hline
\multicolumn{1}{c||}{FashionNet~\cite{DF1}} & \multicolumn{1}{c}{0.0781}          & \multicolumn{1}{c}{0.0803}          & \multicolumn{1}{c}{0.0975}          & \multicolumn{1}{c}{0.0923}          & \multicolumn{1}{c}{0.0874}          & \multicolumn{1}{c}{0.0821}          & \multicolumn{1}{c}{0.0802}          & \multicolumn{1}{c}{0.0893}          & 0.0859          \\
\multicolumn{1}{c||}{DFA~\cite{FLD}}        & \multicolumn{1}{c}{0.0480}          & \multicolumn{1}{c}{0.0480}          & \multicolumn{1}{c}{0.0910}          & \multicolumn{1}{c}{0.0890}          & \multicolumn{1}{c}{-}               & \multicolumn{1}{c}{-}               & \multicolumn{1}{c}{0.0710}          & \multicolumn{1}{c}{0.0720}          & 0.0680          \\ 
\multicolumn{1}{c||}{DLAN~\cite{DLAN}}       & \multicolumn{1}{c}{0.0531}          & \multicolumn{1}{c}{0.0547}          & \multicolumn{1}{c}{0.0705}          & \multicolumn{1}{c}{0.0735}          & \multicolumn{1}{c}{0.0752}          & \multicolumn{1}{c}{0.0748}          & \multicolumn{1}{c}{0.0693}          & \multicolumn{1}{c}{0.0675}          & 0.0672          \\ 
\multicolumn{1}{c||}{BCRNNs~\cite{BCRNN}}     & \multicolumn{1}{c}{0.0463}          & \multicolumn{1}{c}{0.0471}          & \multicolumn{1}{c}{\textbf{0.0627}} & \multicolumn{1}{c}{\textbf{0.0614}} & \multicolumn{1}{c}{0.0635}          & \multicolumn{1}{c}{0.0692}          & \multicolumn{1}{c}{0.0635}          & \multicolumn{1}{c}{\textbf{0.0527}} & 0.0583          \\ \hline
\multicolumn{1}{c||}{Ours}       & \multicolumn{1}{c}{\textbf{0.0386}} & \multicolumn{1}{c}{\textbf{0.0391}} & \multicolumn{1}{c}{0.0675}          & \multicolumn{1}{c}{0.0672}          & \multicolumn{1}{c}{\textbf{0.0576}} & \multicolumn{1}{c}{\textbf{0.0605}} & \multicolumn{1}{c}{\textbf{0.0615}} & \multicolumn{1}{c}{0.0621}          & \textbf{0.0568} \\ \hline \hline
\multicolumn{10}{c}{\textbf{DeepFashion}}\\\hline
\multicolumn{1}{c||}{}           & \multicolumn{1}{c}{L.Collar}        & \multicolumn{1}{c}{R.Collar}        & \multicolumn{1}{c}{L.Sleeve}        & \multicolumn{1}{c}{R.Sleeve}        & \multicolumn{1}{c}{L.Waistline}     & \multicolumn{1}{c}{R.Waistline}     & \multicolumn{1}{c}{L.Hem}           & \multicolumn{1}{c}{R.Hem}           & Avg.            \\ \hline
\multicolumn{1}{c||}{FashionNet~\cite{DF1}} & \multicolumn{1}{c}{0.0854}          & \multicolumn{1}{c}{0.0902}          & \multicolumn{1}{c}{0.0973}          & \multicolumn{1}{c}{0.0935}          & \multicolumn{1}{c}{0.0854}          & \multicolumn{1}{c}{0.0845}          & \multicolumn{1}{c}{0.0812}          & \multicolumn{1}{c}{0.0823}          & 0.0872          \\ 
\multicolumn{1}{c||}{DFA~\cite{FLD}}        & \multicolumn{1}{c}{0.0628}          & \multicolumn{1}{c}{0.0637}          & \multicolumn{1}{c}{0.0658}          & \multicolumn{1}{c}{0.0621}          & \multicolumn{1}{c}{0.0726}          & \multicolumn{1}{c}{0.0702}          & \multicolumn{1}{c}{0.0658}          & \multicolumn{1}{c}{0.0663}          & 0.0660          \\
\multicolumn{1}{c||}{DLAN~\cite{DLAN}}       & \multicolumn{1}{c}{0.0570}          & \multicolumn{1}{c}{0.0611}          & \multicolumn{1}{c}{0.0672}          & \multicolumn{1}{c}{0.0647}          & \multicolumn{1}{c}{0.0703}          & \multicolumn{1}{c}{0.0694}          & \multicolumn{1}{c}{0.0624}          & \multicolumn{1}{c}{0.0627}          & 0.0643          \\
\multicolumn{1}{c||}{BCRNNs~\cite{BCRNN}}     & \multicolumn{1}{c}{0.0415}          & \multicolumn{1}{c}{0.0404}          & \multicolumn{1}{c}{0.0496}          & \multicolumn{1}{c}{0.0449}          & \multicolumn{1}{c}{0.0502}          & \multicolumn{1}{c}{0.0523}          & \multicolumn{1}{c}{0.0537}          & \multicolumn{1}{c}{0.0551}          & 0.0484          \\
\multicolumn{1}{c||}{Liu et al.~\cite{ECCVW}}  & \multicolumn{1}{c}{0.0332}          & \multicolumn{1}{c}{0.0346}          & \multicolumn{1}{c}{0.0487}          & \multicolumn{1}{c}{0.0519}          & \multicolumn{1}{c}{0.0422}          & \multicolumn{1}{c}{0.0429}          & \multicolumn{1}{c}{0.0620}          & \multicolumn{1}{c}{0.0639}          & 0.0474          \\ \hline
\multicolumn{1}{c||}{Ours}       & \multicolumn{1}{c}{\textbf{0.0312}} & \multicolumn{1}{c}{\textbf{0.0324}} & \multicolumn{1}{c}{\textbf{0.0427}} & \multicolumn{1}{c}{\textbf{0.0434}} & \multicolumn{1}{c}{\textbf{0.0361}} & \multicolumn{1}{c}{\textbf{0.0373}} & \multicolumn{1}{c}{\textbf{0.0442}} & \multicolumn{1}{c}{\textbf{0.0475}} & \textbf{0.0393} \\ \hline
\end{tabular}}    
\end{center}
\caption{Quantitative results for clothing landmark detection on Deepfashion and FLD with respect to normalized error (NE). The best results are marked in \textbf{bold}.}
\label{table:results}
\end{table*}
\section{Methods}
\subsection{Network architecture}
Our network consists of three parts: a feature extractor, a global-local embedding module, and an upsampling network.
The overall architecture of the proposed method is illustrated in Fig.~\ref{fig:architecture}.
First, an input fashion image is resized to $224\times224$ and fed into the feature extractor.
For the feature extractor, we use VGG-16 network~\cite{VGG} except the last convolutional layer and initialize weights with parameters pretrained on ImageNet~\cite{imagenet}.
After the conv4\_3 layer of VGG-16, the global-local embedding module is employed to generate rich landmark embedding features.
We describe details of the global-local embedding module in~\ref{non-local module}.

From generated rich features, the landmark localization network predicts landmark heatmaps.
For more accurate landmark estimation, we produce high-resolution landmark heatmaps which have the same size to the input image.
Each value of the heatmaps represents the probability that there is a landmark.
To increase the spatial size of the feature map, we use several transposed convolutions with kernel size 4, stride 2, and padding 1.
At the end of the upsampling network, we utilize a $1\times1$ convolution to produce a heatmap for each landmark.

\subsection{Global-local embedding module} \label{non-local module}
We introduce a global-local embedding module to exploit rich contextual knowledge of a clothing item.
As shown in the bottom of Fig.~\ref{fig:architecture}, the global-local embedding module consists of a non-local block~\cite{NLN} and two convolutional layers.

Let $\mathbf{x}$ denote input features of the global-local embedding module, which is the conv4\_3 feature map in our network.
The output feature is first represented by the non-local block.
Compared to a conventional convolution operation, the non-local operation calculates long-range dependencies between any two different points, and sums up the weighted input features.
This operation is formulated as:
\begin{align}
    \mathbf{y} = \frac{1}{C(\mathbf{x})}f\left(\theta(\mathbf{x}),\phi(\mathbf{x})\right)g(\mathbf{x}),
\end{align}
where $C(\mathbf{x})$ is a normalization factor of $\mathbf{x}$, and $\theta(\cdot)$, $\phi(\cdot)$, and $g(\cdot)$ are $1\times1$ convolutions.
We define the function $f$ as an embedded Gaussian function as follows:
\begin{align}
    f\left(\theta(\mathbf{x}),\phi(\mathbf{x})\right) = \exp \left(\theta(\mathbf{x})^T\phi(\mathbf{x}) \right).
\end{align}
By utilizing the non-local operation with a residual connection, we obtain the $\mathbf{\hat{y}}$:
\begin{align}
    \mathbf{\hat{y}} = w(\mathbf{y}) + \mathbf{x}.
\end{align}
where $w(\cdot)$ is a $1\times1$ convolution.
Finally, $\mathbf{\hat{y}}$ has rich global knowledge of the clothing item by weighted with long-range relationship.

For locating landmark positions, it is necessary to assimilate the global-informative features in a local manner.
To improve the local representation power of the output feature $\mathbf{z}$, two convolutional layers perform a weighted sum in local neighborhoods:
\begin{align}
    \mathbf{z}=F_{2}\left(F_{1}\left(\mathbf{\hat{y}} \right) \right),
\end{align}
where $F_{i}$ denotes the $i$-th convolutional layer including a $3\times3$ convolution, a batch normalization, and a nonlinear function (i.e. ReLU) sequentially.
By this process, the output feature $\mathbf{z}$ is manipulated to have not only global but also local contextual information of the input clothing image.

We use sequentially $k$ global-local embedding modules to gradually produce more advanced deep feature representation.
We empirically set $k$ as 2 for Deepfashion and 3 for FLD.

\begin{figure*}[t]
\begin{center}
\includegraphics[width=170mm]{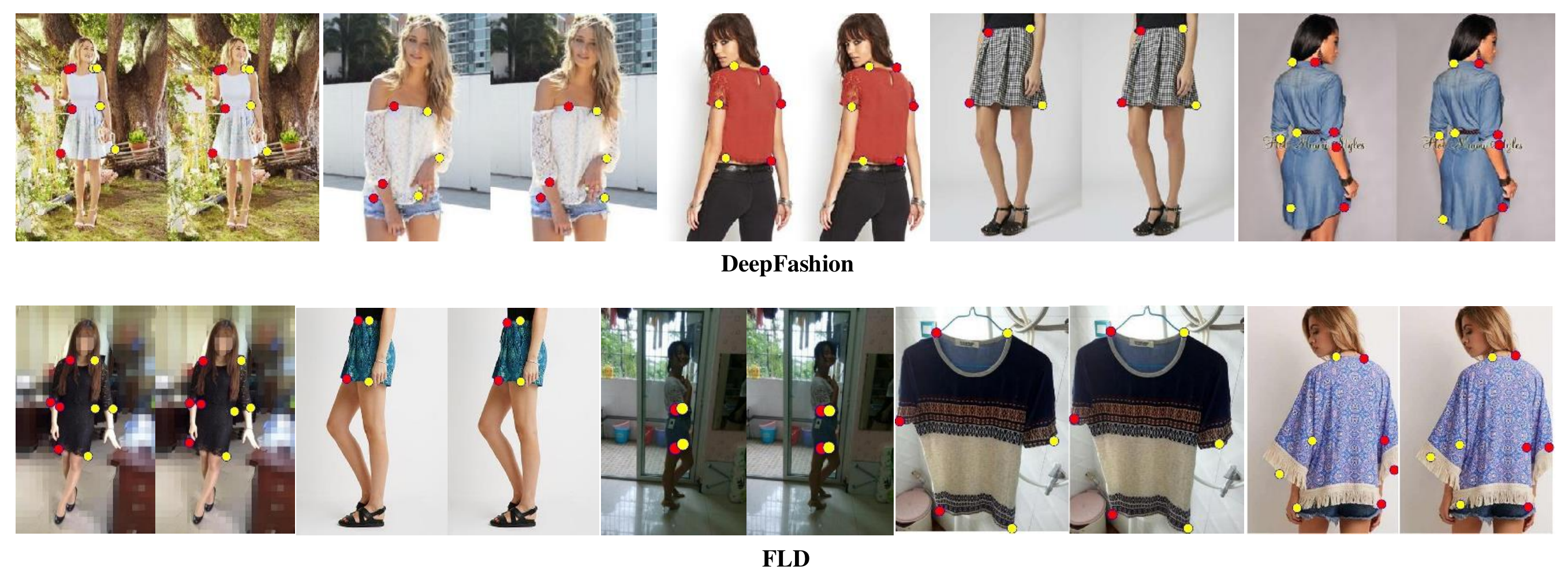}
\end{center}
  \caption{Qualitative results on Deepfashion (first row) and FLD (second row). For each example, ground-truth landmarks are shown in right image and predicted landmarks are shown in left image. Red and yellow circles indicate right-side and left-side landmarks, respectively.}
\label{fig:result}
\end{figure*}
\section{Experiments}
\subsection{The benchmark datasets and evaluation}
We evaluate the proposed network on two datasets, DeepFashion~\cite{DF1} and FLD~\cite{FLD}.
DeepFashion is a large fashion dataset with 289,222 images. 
Those images are composed of 209,222 images for training, 40,000 images for validation, and 40,000 images for testing.
FLD~\cite{FLD} is a fashion landmark dataset with 123,016 images with more diverse variations in poses, scales, and background.
FLD images are divided into 83,033 images for training, 19,991 images for validation, and remaining 19,991 images for testing.
Each image of both two datasets is annotated with bounding boxes and landmarks.
There are 8 landmarks for full-body clothes, 6 landmarks for upper clothes, and 4 landmarks for lower clothes.

We adopt normalized error (NE) metric~\cite{FLD} for evaluation of our proposed method and comparisons to the other state-of-the-art methods.
NE is the $\mathit{L}_2$ distance between predicted and ground-truth landmarks in the normalized coordinate space, formulated as:
\begin{align}
\textrm{NE} = \frac{1}{N}\sum_{i=1}^{N} \frac{\begin{Vmatrix}
p_{i}-\hat{p}_{i}
\end{Vmatrix}}{h_{i} \times w_{i}},
\end{align}
where $p_{i}$ and $\hat{p}_{i}$ are ground-truth and predicted landmarks of $i$-th sample respectively, $N$ is the number of samples, and $h_{i}$ and $w_{i}$ are the height and width of $i$-the sample.
\subsection{Results}
We conduct experiments on two large datasets and compare results of the proposed network with the state-of-the-art methods~\cite{DF1,FLD,DLAN,BCRNN,ECCVW}. 
Table~\ref{table:results} summaries comparison results.
From the table, our proposed network outperforms all the competitors at 0.0568 on FLD and 0.0393 on DeepFashion.
Among the landmarks, it is hard to discriminate between waistlines and hems landmarks which have the largest error rate in other methods.
Our proposed method achieves performance improvements in detecting waistlines and hems landmarks.
From these results, it is proved that our network learns more advanced feature representations for fashion landmark detection with the aid of the proposed global-local embedding module.

We also visualize landmark detection results in Fig.~\ref{fig:result}.
We can see that the proposed model discriminates right- and left-side landmarks even in the back-view and side-view images.
According to these results, it is demonstrated that our network is robust to view-point variations and deformations.

\section{Conclusion}
In this paper, we have proposed a fashion landmark detection network with a global-local embedding module.
Our proposed global-local embedding module is based on a non-local operation and a convolution operation.
Utilizing global-local embedding module facilitated to exploit not only global but also local contextual knowledge of a clothing item.
We evaluated our method on two benchmark datasets, and achieved the state-of-the-art performance over recent methods.
Experimental results demonstrated that our proposed method improves the feature representation of a clothing item for fashion landmark detection.

\section*{Acknowledgement}\emph{•}
This work was supported by Electronics and Telecommunications Research Institute(ETRI) grant funded by the Korean government.
[19ZS1100, Core Technology Research for Self-Improving Artificial Intelligence System]
{\small
\bibliographystyle{ieee_fullname}
\bibliography{egpaper}

\begin{thebibliography}{10}\itemsep=-1pt

\bibitem{DF2}
Y. Ge, R. Zhang, X. Wang, X. Tang, and P. Luo.
\newblock Deepfashion2: A versatile benchmark for detection, pose estimation,
  segmentation and re-identification of clothing images.
\newblock In {\em IEEE Conference on Computer Vision and Pattern Recognition
  (CVPR)}. 2019.

\bibitem{WTBI}
K.~M. Hadi, X. Han, S. Lazebnik, A.~C. Berg, and T.~L. Berg.
\newblock Where to buy it: Matching street clothing photos in online shops.
\newblock In {\em IEEE International Conference on Computer Vision (ICCV)}.
  2015.

\bibitem{generation2}
X. Han, Z. Wu, Z. Wu, R. Yu, and L.~S. Davis.
\newblock Viton: An image-based virtual try-on network.
\newblock In {\em IEEE Conference on Computer Vision and Pattern Recognition
  (CVPR)}. 2018.

\bibitem{retrieval}
J. Huang, R.~S. Feris, Q. Chen, and S. Yan.
\newblock Cross-domain image retrieval with a dual attribute-aware ranking
  network.
\newblock In {\em IEEE International Conference on Computer Vision (ICCV)}.
  2015.

\bibitem{imagenet}
A. Krizhevsky, I. Sutskever, and G.~E. Hinton.
\newblock Imagenet classification with deep convolutional neural networks.
\newblock In {\em Advances in Neural Information Processing Systems (NIPS)}.
  2012.

\bibitem{ECCVW}
J. Liu and H. Lu.
\newblock Deep fashion analysis with feature map upsampling and landmark-driven
  attention.
\newblock In {\em European Conference on Computer Vision (ECCV) Workshop}.
  Springer, 2018.

\bibitem{DF1}
Z. Liu, P. Luo, S. Qiu, X. Wang, and X. Tang.
\newblock Deepfashion: Powering robust clothes recognition and retrieval with
  rich annotations.
\newblock In {\em IEEE Conference on Computer Vision and Pattern Recognition
  (CVPR)}. 2016.

\bibitem{FLD}
Z. Liu, S. Yan, P. Luo, X. Wang, and X. Tang.
\newblock Fashion landmark detection in the wild.
\newblock In {\em European Conference on Computer Vision (ECCV)}. Springer,
  2016.

\bibitem{VGG}
K. Simonyan and A. Zisserman.
\newblock Very deep convolutional networks for large-scale image recognition.
\newblock In {\em International Conference on Learning Representations (ICLR)}.
  2015.

\bibitem{BCRNN}
W. Wang, Y. Xu, J. Shen, and S. Zhu.
\newblock Attentive fashion grammar network for fashion landmark detection and
  clothing category classification.
\newblock In {\em IEEE Conference on Computer Vision and Pattern Recognition
  (CVPR)}. 2018.

\bibitem{NLN}
X. Wang, R. Girshick, A. Gupta, and K. He.
\newblock Non-local neural networks.
\newblock In {\em IEEE Conference on Computer Vision and Pattern Recognition
  (CVPR)}. 2018.

\bibitem{DLAN}
S. Yan, Z. Liu, P. Luo, S. Qiu, X. Wang, and X. Tang.
\newblock Unconstrained fashion landmark detection via hierarchical recurrent
  transformer networks.
\newblock In {\em ACM International Conference on Multimedia}. 2017.

\bibitem{generation}
S. Zhu, R. Urtasun, S. Fidler, D. Lin, and C. Change~Loy.
\newblock Be your own prada: Fashion synthesis with structural coherence.
\newblock In {\em IEEE International Conference on Computer Vision (ICCV)}.
  2017.

\end{thebibliography}
}

\end{document}